\documentclass[11pt,a4paper]{article}
\usepackage[hyperref]{naaclhlt2018}
\usepackage{times}
\usepackage{latexsym}

\usepackage{url}

\usepackage{amsmath}
\usepackage{dcolumn}
\usepackage{hhline}
\usepackage{verbatim}
\usepackage{multirow}
\usepackage{array}
\usepackage{graphicx}
\usepackage{kotex}
\usepackage{amssymb}
\usepackage{rotating}

\aclfinalcopy 
\def\aclpaperid{348} 

\newcolumntype{L}[1]{>{\raggedright\let\newline\\\arraybackslash\hspace{0pt}}m{#1}}
\newcolumntype{C}[1]{>{\centering\let\newline\\\arraybackslash\hspace{0pt}}m{#1}}
\newcolumntype{R}[1]{>{\raggedleft\let\newline\\\arraybackslash\hspace{0pt}}m{#1}}

\newcommand\Tstrut{\rule{0pt}{2.0ex}}         

\title{Learning to Rank Question-Answer Pairs using \\ Hierarchical Recurrent Encoder with Latent Topic Clustering}

\author{Seunghyun Yoon,~Joongbo Shin~and~Kyomin Jung \\
  Dept. of Electrical and Computer Engineering \\
  Seoul National University, Seoul, Korea \\
  {\tt \{mysmilesh,jbshin,kjung\}@snu.ac.kr} }

\date{}

\begin{document}
\maketitle

\begin{abstract}
In this paper, we propose a novel end-to-end neural architecture for ranking candidate answers, that adapts a hierarchical recurrent neural network and a latent topic clustering module. 
With our proposed model, a text is encoded to a vector representation from an word-level to a \textit{chunk-level} to effectively capture the entire meaning. 
In particular, by adapting the hierarchical structure, our model shows very small performance degradations in longer text comprehension while other state-of-the-art recurrent neural network models suffer from it.
Additionally, the latent topic clustering module extracts semantic information from target samples.
This clustering module is useful for any text related tasks by allowing each data sample to find its nearest topic cluster, thus helping the neural network model analyze the entire data. 
We evaluate our models on the Ubuntu Dialogue Corpus and consumer electronic domain question answering dataset, which is related to Samsung products. 
The proposed model shows state-of-the-art results for ranking question-answer pairs.
\end{abstract}

\section{Introduction}
\label{intro}
Recently neural network architectures have shown great success in many machine learning fields such as image classification, speech recognition, machine translation, chat-bot, question answering, and other task-oriented areas. Among these, the automatic question answering (QA) task has long been considered a primary objective of artificial intelligence.

In the commercial sphere, the QA task is usually tackled by using pre-organized knowledge bases and/or by using information retrieval (IR) based methods, which are applied in popular intelligent voice agents such as \textit{Siri}, \textit{Alexa}, and \textit{Google Assistant} (from Apple, Amazon, and Google, respectively).
Another type of advanced QA systems is IBM's Watson who builds knowledge bases from unstructured data. These raw data are also indexed in search clusters to support user queries \cite{fan2012automatic,chu2012finding}. 



In academic literature, researchers have intensely studied sentence pair ranking task which is core technique in QA system. The ranking task selects the best answer among candidates retrieved from knowledge bases or IR based modules. 
Many neural network architectures with end-to-end learning methods are proposed to address this task \cite{yin2016abcnn, wang2016compare,wang2017bilateral}.
These works focus on matching sentence-level text pair \cite{wang2007jeopardy,yang2015wikiqa,bowman2015large}. Therefore, they have limitations in understanding longer text such as multi-turn dialogue and explanatory document, resulting in performance degradation on ranking as the length of the text become longer.

With the advent of the huge multi-turn dialogue corpus \cite{lowe2015ubuntu}, researchers have proposed neural network models to rank longer text pair \cite{kadlec2015improved,baudivs2016sentence}.
These techniques are essential for capturing context information in multi-turn conversation or understanding multiple sentences in explanatory text.


In this paper, we focus on investigating a novel neural network architecture with additional data clustering module to improve the performance in ranking answer candidates which are longer than a single sentence.
This work can be used not only for the QA ranking task, but also to evaluate the relevance of next utterance with given dialogue generated from the dialogue model. 
The key contributions of our work are as follows:

First, we introduce a Hierarchical Recurrent Dual Encoder (HRDE) model to effectively calculate the affinity among question-answer pairs to determine the ranking.
By encoding texts from an word-level to a chunk-level with hierarchical architecture, the HRDE prevents performance degradations in understanding longer texts while other state-of-the-art neural network models suffer.

Second, we propose a Latent Topic Clustering (LTC) module to extract latent information from the target dataset, and apply these additional information in end-to-end training.
This module allows each data sample to find its nearest topic cluster, thus helping the neural network model analyze the entire data. 
The LTC module can be combined to any neural network as a source of additional information. 
This is a novel approach using latent topic cluster information for the QA task, especially by applying the combined model of HRDE and LTC to the QA pair ranking task. 

Extensive experiments are conducted to investigate efficacy and properties of the proposed model.
Our proposed model outperforms previous state-of-the-art methods in the Ubuntu Dialogue Corpus, which is one of the largest text pair scoring datasets.
We also evaluate the model on real world QA data crawled from crowd-QA web pages and from Samsung's official web pages. 
Our model also shows the best results for the QA data when compared to previous neural network based models.

\section{Related Work}
\label{realtedword}
Researchers have released question and answer datasets for research purposes and have proposed various models to solve these datasets. \cite{wang2007jeopardy,yang2015wikiqa,tan2015lstm} introduced small dataset to rank sentences that have higher probabilities of answering questions such as  WikiQA and insuranceQA.
To alleviate the difficulty in aggregating datasets, that are large and have no license restrictions, some researchers introduced new datasets for sentence similarity rankings \cite{baudivs2016sentence,lowe2015ubuntu}. As of now, the Ubuntu Dialogue dataset is one of the largest corpus openly available for text ranking.

To tackle the Ubuntu dataset, \cite{lowe2015ubuntu} adopted the ``term frequency-inverse document frequency" approach to capture important words among context and next utterances \cite{ramos2003using}. \cite{bordes2014open,yu2014deep} proposed deep neural network architecture for embedding sentences and measuring similarities to select answer sentence for a given question. \cite{kadlec2015improved} used convolution neural network (CNN) architecture to embed the sentence while a final output vector was compared to the target text to calculate the matching score. They also tried using long short-term memory (LSTM) \cite{hochreiter1997long}, bi-directional LSTM and ensemble method with all of those neural network architectures and achieved the best results on the Ubuntu Dialogues Corpus dataset. 
Another type of neural architecture is the RNN-CNN model, which encodes each token with a recurrent neural network (RNN) and then feeds them to the CNN \cite{baudivs2016sentence}. 
Researchers also introduced an attention based model to improve the performance~\cite{tan2015lstm,wang2016compare,wang2017bilateral}. 

Recently, the hierarchical recurrent encoder-decoder model was proposed to embed contextual information in user query prediction and dialogue generation tasks \cite{sordoni2015hierarchical,serban2016building}.
This shows improvement in the dialogue generation model where the context for the utterance is important.
As another type of neural network architecture, memory network was proposed by \cite{sukhbaatar2015end}. Several researchers adopted this architecture for the reading comprehension (RC) style QA tasks, because it can extract contextual information from each sentence and use it in finding the answer \cite{xiong2016dynamic,kumar2016ask}.  However, none of this research is applied to the QA pair ranking task directly.

\section{Model}
\label{model}
In this section, we depict a previously released neural text ranking model, and then introduce our proposed neural network model.

\subsection{Recurrent Dual Encoder (RDE)}
A subset of sequential data is fed into the recurrent neural network (RNN) which leads to the formation of the network's internal hidden state $h_{t}$ to model the time series patterns. This internal hidden state is updated at each time step with the input data $w_{t}$ and the hidden state of the previous time step $h_{t-1}$ as follows:  
\begin{equation}
\begin{aligned}
& h_t = f_{\theta}(h_{t-1}, w_t),
\end{aligned}
\end{equation}
where $f_{\theta}$ is the RNN function with weight parameter $\theta$, $h_t$ is hidden state at \textit{t}-th word input, $w_t$ is \textit{t}-th word in a target question $\boldsymbol{w}^Q={\{w^Q_{1:t_q}\}}$ or an answer text $\boldsymbol{w}^A={\{w^A_{1:t_a}\}}$ . 

The previous RDE model uses two RNNs for encoding question text and answer text to calculate affinity among texts \cite{lowe2015ubuntu}.
After encoding each part of the data, the affinity among the text pairs is calculated by using the final hidden state value of each question and answer RNNs. 
The matching probability between question text $\boldsymbol{w}^Q$ and answer text $\boldsymbol{w}^A$ with the training objective are as follows:
\begin{equation}
\begin{aligned}
& p(\text{label}) = \sigma ( ({h_{t_q}^Q})^T M~h_{t_a}^A + b ), \\
& \mathcal{L} = -\log \prod_{n=1}^{N} p(\text{label}_n | h_{n,t_q}^Q, h_{n,t_a}^A),
\end{aligned}
\label{eq_de_loss}
\end{equation}
where $h_{t_q}^Q$ and $h_{t_a}^A$ are last hidden state of each question and answer RNN with the dimensionality $h_{t} \in \mathbb{R}^d$. The $M \in \mathbb{R}^{d \times d}$ and bias $b$ are learned model parameters. The $N$ is total number of samples used in training and $\sigma$ is the sigmoid function. 

\begin{figure}[t]
\centering
\includegraphics[width=1\columnwidth]{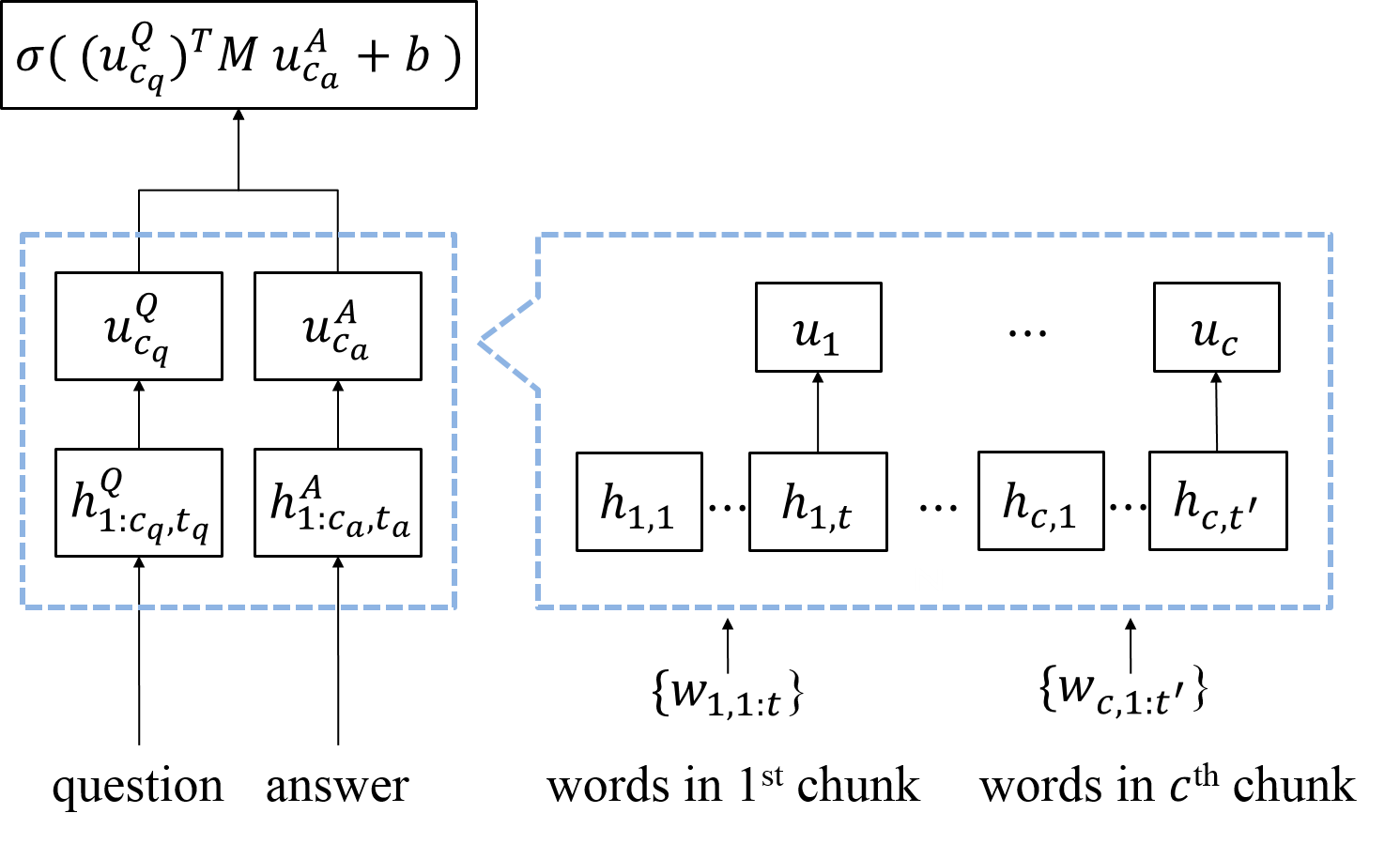}
\caption{
Diagram of the HRDE model.
The word-lever RNN encodes words sequences of each chunk. The the final hidden status of the word-level RNN is fed into chunk-level RNN.
}
\label{fig_hrde}
\end{figure}

\subsection{Hierarchical Recurrent Dual Encoder (HRDE)}
From now we explain our proposed model. 
The previous RDE model tries to encode the text in question or in answer with RNN architecture. It would be less effective as the length of the word sequences in the text increases because RNN's natural characteristic of forgetting information from long ranging data. 
To address this RNN's forgetting phenomenon, \cite{bahdanau2014neural} proposed an attention mechanism, however, we found that it still showed a limitation when we consider very large sequential length data such as 162 steps average in the Ubuntu Dialogue Corpus dataset (see Table~\ref{t_data_stat}). 
To overcome this limitation, we designed the HRDE architecture. The HRDE model divides long sequential text data into small chunk such as sentences, and encodes the whole text from word-level to chunk-level by using two hierarchical level of RNN architecture. 

Figure \ref{fig_hrde} shows a diagram of the HRDE model. The word-level RNN part is responsible for encoding the words sequence 
$\boldsymbol{w}_c={\{w_{c,1:t}\}}$ in each chunk.
The chunk can be sentences in paragraph, paragraphs in essay, turns in dialogue or any kinds of smaller meaningful sub-set from the text. Then the final hidden states of each chunk will be fed into chunk-level RNN with its original sequence order kept. Therefore the chunk-level RNN can deal with pre-encoded chunk data with less sequential steps. The hidden states of the hierarchical RNNs are as follows:
\begin{equation}
\begin{aligned}
 &h_{c,t} = f_{\theta}(h_{c,t-1}, w_{c,t}), \\
 &u_c = g_{\theta}(u_{c-1}, h_c),
\end{aligned}
\label{eq_hrde}
\end{equation}
where $f_\theta$ and $g_\theta$ are the RNN function in hierarchical architecture with weight parameters $\theta$, $h_{c,t}$ is word-level RNN's hidden status at \textit{t}-th word in \textit{c}-th chunk. The $w_{c,t}$ is \textit{t}-th word in \textit{c}-th chunk of target question or answer text. The $u_c$ is chunk-level RNN's hidden state at \textit{c}-th chunk sequence, and $h_c$ is word-level RNN's last hidden state of each chunk $h_c \in\{h_{1:c,t}\}$.  

We use the same training objective as the RDE model, and the final matching probability between question and answer text is calculated using  chunk-level RNN as follows:
\begin{equation}
\begin{aligned}
& p(\mbox{label}) = \sigma ( ({u_{c_q}^Q})^T M~u_{c_a}^A + b ),
\end{aligned}
\label{eq_hrde_loss}
\end{equation}
where $u_{c_q}^Q$ and $u_{c_a}^A$ are chunk-level RNN's last hidden state of each question and answer text with the dimensionality $u_c \in \mathbb{R}^{d^{u}}$, which involves the $M \in \mathbb{R}^{d^{u} \times d^{u}}$.

\begin{figure}[t]
\centering
\includegraphics[width=0.90\columnwidth]{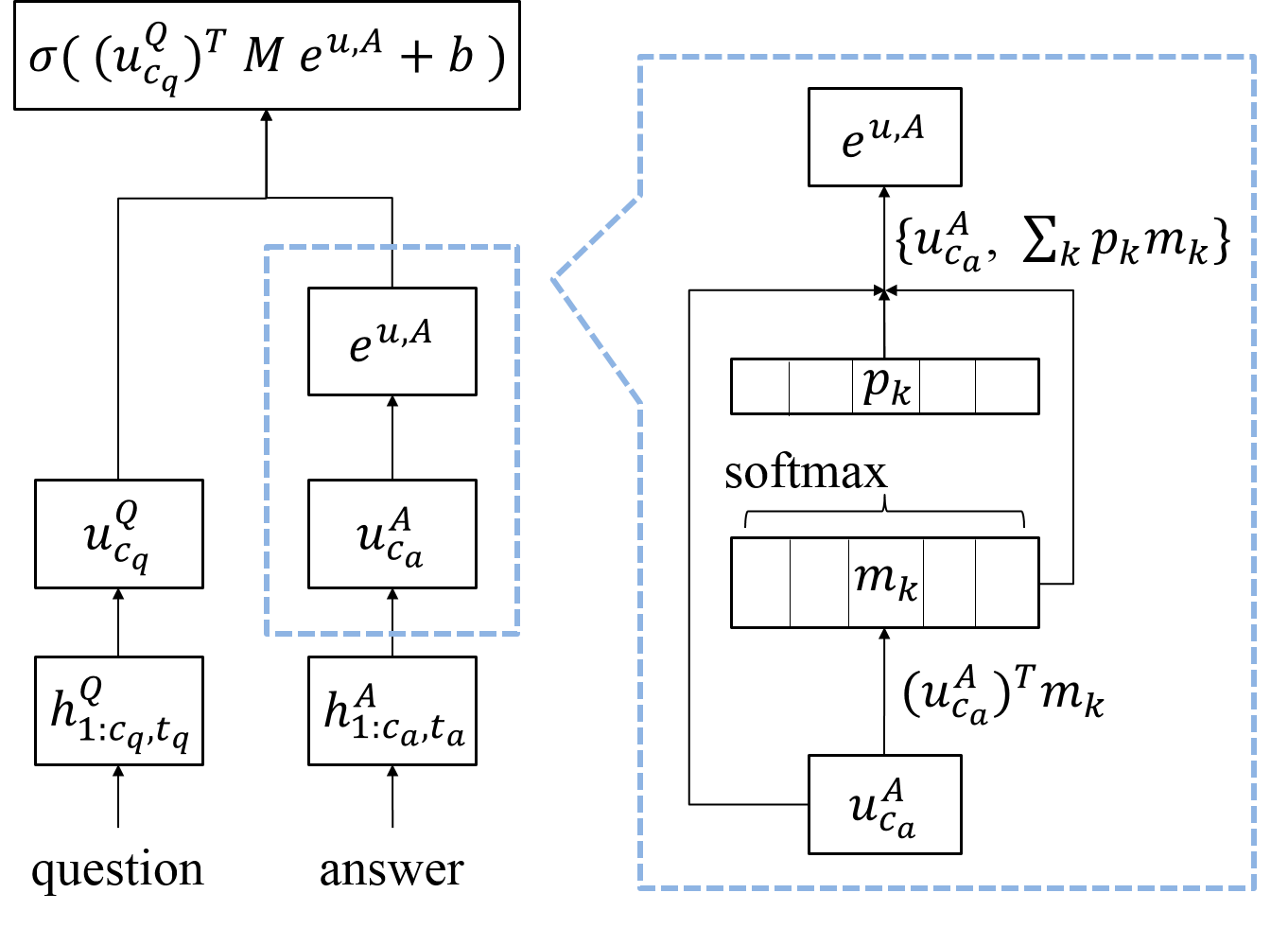}
\caption{
Diagram of the HRDE-LTC.
Input vector is compared to each latent topic memory $m_k$ to calculate cluster-info contained vector. This vector will be concatenated to original input vector.
}
\label{fig_latent_topic}
\end{figure}

\subsection{Latent Topic Clustering (LTC)}
To learn how to rank QA pairs, a neural network should be trained to find the proper feature that represents the information within the data and fits the model parameter that can approximate the true-hypothesis. For this type of problem, we propose the LTC module for grouping the target data to help the neural network find the true-hypothesis with more information from the topic cluster in end-to-end training. 

The blue-dotted box on the right-side of Figure \ref{fig_latent_topic} shows LTC structure diagram. To assign topic information, we build internal latent topic memory $m \in \mathbb{R}^{d^{m} \times K}$, which is only model parameter to be learned, where $d^{m}$ is vector dimension of each latent topic and $K$ is number of latent topic cluster.
For a given input sequence  $\boldsymbol{x}={\{x_{1:t}\}}$with these K vectors, we construct LTC process as follows:
\begin{equation}
\begin{aligned}
& p_k = \text{softmax}(( \boldsymbol{x})^Tm_k), \\
& \boldsymbol{x}_K = \sum_{k=1}^{K} p_km_k, \\
& \boldsymbol{e} = \text{concat}\{\boldsymbol{x},\boldsymbol{x}_K\}.
\end{aligned}
\label{eq_ltc_02}
\end{equation}

First, the similarity between the $\boldsymbol{x}$ and each latent topic vector is calculated by dot-product.
Then the resulting $K$ values are  normalized by the softmax function $\text{softmax}(z_k)=e^{z_k}/ \sum_i e^{z_i}$ to produce a similarity probability $p_k$.
After calculating the latent topic probability $p_k$, $x_K$ is retrieved from summing over $m_k$ weighted by the $p_k$. Then we concatenate this result with the original encoding vector to generate the final encoding vector 
$\boldsymbol{e}$ with the LTC information added.

Note that the input sequence of the LTC could be any type of neural network based encoding function $\boldsymbol{x}=f^{\text{enc}}_\theta(\boldsymbol{w})$ such as RNN, CNN and multilayer perceptron model (MLP).
In addition, if the dimension size of $\boldsymbol{x}$ is different from that of memory vector, additional output projection layer should be placed after $\boldsymbol{x}$ before applying dot-product to the memory.

\subsection{Combined Model of (H)RDE and LTC}
As the LTC module extracts additional topic cluster information from the input data, we can combine this module with any neural network in their end-to-end training flow. In our experiments, we combine the LTC module with the RDE and HRDE models.

\subsubsection{RDE with LTC}
The RDE model encodes question and answer texts to $h_{t_q}^Q$ and $h_{t_a}^A$, respectively. Hence, the LTC module could take these vectors as the input to generate latent topic cluster information added vector $\boldsymbol{e}$. 
With this vector, we calculate the affinity among question and answer texts as well as additional cluster information.
The following equation shows our RDE-LTC process:
\begin{equation}
\begin{aligned}
& p(\text{label}) = \sigma ( ({h_{t_q}^Q})^T M~e^A + b ). \\
\end{aligned}
\label{eq_rde_ltc}
\end{equation}
In this case, we applied the LTC module only for the answer side, assuming that the answer text is longer than the question. Thus, it needs to be clustered. 
To train the network, we use the same training objective, to minimize cross-entropy loss, as in equation (\ref{eq_de_loss}).

\subsubsection{HRDE with LTC}
The LTC can be combined with the HRDE model, in the same way it is applied to the RDE-LTC model by modifying equation (\ref{eq_rde_ltc} as follows:
\begin{equation}
\begin{aligned}
& p(\text{label}) = \sigma ( ({u_{c_q}^Q})^T M~e^{u,A} + b ),
\end{aligned}
\end{equation}
where $u_{c_q}^Q$  is the final network hidden state vector of the chunk-level RNN for a question input sequence. The $e^{u,A}$ is the LTC information added vector from equation (\ref{eq_ltc_02}), where the LTC module takes the input $\boldsymbol{x}=\boldsymbol{u}^A$ from the HRDE model equation (\ref{eq_hrde}).
The HRDE-LTC model also use  the  same training objective, minimizing cross-entropy loss, as in equation (\ref{eq_de_loss}).
Figure \ref{fig_latent_topic} shows a diagram of the combined model with the HRDE and the LTC.

\begin{table*}[t]
\small
\centering
\begin{tabular}{|c|c|c|c|c|c|c|c|c|c|}
\hline
  \multirow{2}{*}{ \textbf{Dataset}} & \multicolumn{3}{c|}{ \textbf{\# Samples} } & \multicolumn{3}{c|}{ \textbf{Message (Avg.)} } & \multicolumn{3}{c|}{ \textbf{Response (Avg.)} } \\ \cline{2-10} 
& Train     & Val.      & Test     & \begin{tabular}[c]{@{}c@{}} \# tokens\end{tabular}  & \begin{tabular}[c]{@{}c@{}} \# groups\end{tabular}   & \begin{tabular}[c]{@{}c@{}} \# tokens\\ /group\end{tabular} & \begin{tabular}[c]{@{}c@{}} \# tokens\end{tabular}    & \begin{tabular}[c]{@{}c@{}} \# groups\end{tabular} & \begin{tabular}[c]{@{}c@{}} \# tokens\\ /group\end{tabular} \\ 
\hhline{|==========|}

Ubuntu-v1\Tstrut  & 1M\Tstrut   & 35,609\Tstrut   & 35,517\Tstrut   & $\underset{~~\pm 132.47}{162.47}$\Tstrut   & $\underset{~~\pm 6.32}{8.43}$\Tstrut   & $\underset{~~\pm 18.41}{20.14}$\Tstrut   & $\underset{~~\pm 13.93}{14.44}$\Tstrut   & 1  & -\Tstrut  \\ \hline

Ubuntu-v2  & 1M  & 19,560  & 18,920  & $\underset{~~\pm 74.71}{85.92}$  & $\underset{~~\pm 2.98}{4.95}$  & $\underset{~~\pm 20.19}{20.73}$  & $\underset{~~\pm 16.41}{17.01}$  & 1  & -\Tstrut  \\ \hline

Samsung QA  & 163,616   & 10,000   & 10,000   & $\underset{~~\pm 6.42}{12.84}$ & 1  & -  & $\underset{~~\pm 192.12}{173.48}$  & $\underset{~~\pm 5.58}{6.09}$  & $\underset{~~\pm 31.91}{29.28}$\Tstrut  \\ \hline

\end{tabular}
\caption{Properties of the Ubuntu and Samsung QA dataset. The message and response are \{context\}, \{response\} in Ubuntu and \{question\}, \{answer\} in the Samsung QA dataset. 
}
\label{t_data_stat}
\end{table*}
\begin{table}[bh!]
\centering

\begin{tabular}{|L{0.93\columnwidth}|}
\hline
\textbf{Question} \Tstrut \\ 
\hline
how do i set a timer of clock in applications and development for samsung galaxy s4 mini?\Tstrut\\
\hline
\end{tabular}

\vskip 2mm

\begin{tabular}{|L{0.93\columnwidth}|}
\hline
\textbf{Answer} \Tstrut \\ 
\hline
1 from within the clock application, tap timer tab. 2 tap the hours, minutes, or seconds field and use the on-screen keypad to enter the hour, minute, or seconds. the timer plays an alarm at the end of the countdown. 3 tap start to start the timer. 4 tap stop to stop the timer or reset to reset the timer and start over. 5 tap restart to resume the timer counter. \Tstrut \\
\hline
\end{tabular}

\caption{Example of the Samsung QA dataset.}
\label{t_sample_samsungQA}
\end{table}

\section{Experimental Setup and Dataset}
\label{dataset}
\subsection{The Ubuntu Dialogue Corpus}
The Ubuntu Dialogue Corpus has been developed by expanding and preprocessing the Ubuntu Chat Logs\footnote{These logs are available from  http://irclogs.ubuntu.com}, which refer to a collection of logs from the Ubuntu-related chat room for solving problem in using the Ubuntu system by \cite{lowe2015ubuntu}.

Among the utterances in the dialogues, they consider each utterance, starting from the third one, as a potential \{response\} while the previous utterance is considered as a \{context\}. 
The data was processed extracting (\{context\}, \{response\}, flag) tuples from  the dialogues. 

We called this original Ubuntu dataset as Ubuntu-v1 dataset.
After releasing the Ubuntu-v1 dataset,
researchers published v2 version of this dataset.
Main updates are separating train/valid/test dataset by time so that mimics real life implementation, where we are training a model on past data to predict future data, changing sampling procedure to increase average turns in the \{context\}.
We consider this Ubuntu dataset is one of the best dataset in terms of its quality, quantity and availability for evaluating the performance of the text ranking model. 


To encode the text with the HRDE and HRDE-LTC model, a text needs to be divided into several chunk sequences with predefined criteria.
For the Ubuntu-v1 dataset case, we divide the \{context\} part by splitting with  end-of-sentence delimiter ``\_eos\_", and we do not split the \{response\} part since it is normally short and does not contain ``\_eos\_" information. For the Ubuntu-v2 dataset case, we split the \{context\} part in the same way as we do in the Ubuntu-v1 dataset while only using end-of-turn delimiter ``\_eot\_". Table \ref{t_data_stat} shows properties of the Ubuntu dataset.

\subsection{Consumer Product QA Corpus}
To test the robustness of the proposed model, we introduce an additional question and answer pair dataset related to an actual user's interaction with the consumer electronic product domain. 
We crawled data from various sources like the Samsung Electronics' official web site\footnote{http://www.samsung.com/us} and crowd QA web sites\footnote{https://answers.yahoo.com,~http://answers.us.samsung.com} in a similar way that ~\cite{yoon2016automatic} did in building QA system for consumer products.
On the official web page, we can retrieve data consisting of user questions and matched answers like frequently asked questions and troubleshooting.
From the crowd QA sites, there are many answers from various users for each question. 
Among these answers, we choose answers from company certificated users to keep the reliability of the answers high.
If there are no such answers, we skip that question answer pair. 
Table~\ref{t_sample_samsungQA} shows an example of question-answer pair crawled from the web page.
In addition, we crawl hierarchical product category information related to QA pairs. In particular, \textit{mobile}, \textit{office}, \textit{photo}, \textit{tv/video}, \textit{accessories}, and \textit{home appliance} as top-level categories, and specific categories like \textit{galaxy s7}, \textit{tablet}, \textit{led tv}, and \textit{others} are used. We collected these meta-information for further use. 
The total size of the Samsung QA data is over 100,000 pairs and we split the data into approximately 80,000/10,000/10,000 samples to create train/valid/test sets, respectively. 
To create the train set, we use a QA pair sample as a ground-truth and perform negative sampling for answers among training sets to create \textit{false}-label datasets.  In this way, we generated (\{question\}, \{answer\}, flag) triples  (see Table \ref{t_data_stat}). 
We do the same procedure to create valid and test sets by only differentiating more negative sampling within each dataset to generate 9 \textit{false}-label samples with one ground-truth sample. We apply the same method in such a way that the Ubuntu dataset is generated from the Ubuntu Dialogue Corpus to maintain the consistency.
The Samsung QA dataset is available via web repository.
We refer the readers to Appendix~\ref{supplemental} for more examples of each dataset.

\subsection{Implementation Details}
\subsubsection{Ubuntu dataset case} To implement the RDE model, we use two single layer Gated Recurrent Unit (GRU) \cite{chung2014empirical} with 300 hidden units .
Each GRU is used to encode \{context\} and \{response\}, respectively. The weight for the two GRU are shared. The hidden units weight matrix of the GRU are initialized using orthogonal weights \cite{saxe2013exact}, while input embedding weight matrix is initialized using a pre-trained embedding vector, the Glove \cite{pennington2014glove}, with 300 dimension. The vocabulary size is 144,953 and 183,045 for the Ubuntu-v1/v2 case, respectively.
We use the Adam optimizer \cite{kingma2014adam}, with gradients clipped with norm value 1. The maximum time step for calculating gradient of the RNN is determined according to the input data statistics in Table \ref{t_data_stat}. 

For the HRDE model, we use two single layer GRU with 300 hidden units for word-level RNN part, and another two single layer GRU with 300 hidden units for chunk-level RNN part. The weight of the GRU is shared within the same hierarchical part, word-level and chunk-level. The other settings are the same with the RDE model case. As for the combined model with the (H)RDE and the LTC, we choose the latent topic memory dimensions as 256 in both ubuntu-v1 and ubuntu-v2. The number of the cluster in LTC module is decided to 3 for both the RDE-LTC and the HRDE-LTC cases. In HRDE-LTC case, we applied LTC module to the \{context\} part because  we think it is longer having enough information to be clustered with.
All of these hyper-parameters are selected from additional parameter searching experiments.

The dropout \cite{srivastava2014dropout} is applied for the purpose of  regularization with the ratio of: 0.2 for the RNN in the RDE and the RDE-LTC, 0.3 for the word-level RNN part in the HRDE and the HRDE-LTC, 0.8 for the latent topic memory in the RDE-LTC and the HRDE-LTC.

We need to mention that our implementation of the RDE module has the same architecture as the LSTM model \cite{kadlec2015improved} in ubuntu-v1/v2 experiments case. It is also the same architecture with the RNN model \cite{baudivs2016sentence} in ubuntu-v2 experiment case. 
We implement the same model ourselves, because we need a baseline model to compare with other proposed models such as the RDE-LTC, HRDE and HRDE-LTC.

\subsubsection{Samsung QA dataset case} To test the Samsung QA dataset, we use the same implementation of the model (RDE, RDE-LTC, HRDE and HRDE-LTC) used in testing the Ubuntu dataset. Only the differences are, we use 100 hidden units for the RDE and the RDE-LTC, 300 hidden units for the HRDE and 200 hidden units for the HRDE-LTC, and the vocabulary size of 28,848. As for the combined model with the (H)RDE and LTC, the dimensions of the latent topic memory is 64 and the number of latent cluster is 4. We chose best performing hyper-parameter of each model by additional extensive hyper-parameter search experiments.

All of the code developed for the empirical results are available via web repository
\footnote{http://github.com/david-yoon/QA\_HRDE\_LTC}.

\section{Empirical Results}
\label{experiments}
\subsection{Evaluation Metrics}
We regards all the tasks as selecting the best answer among text candidates for the given question. Following the previous work \cite{lowe2015ubuntu}, we report model performance as recall at $k$ (R@k) relevant texts among given 2 or 10 candidates (e.g., 1 in 2 R@1). Though this metric is useful for ranking task, R@1 metric is also meaningful for classifying the best relevant text.

Each model we implement is trained multiple times (10 and 15 times for Ubuntu and the Samsung QA datasets in our experiments, respectively) with random weight initialization, which largely influences performance of neural network model. Hence we report model performance as mean and standard derivation values (Mean$\pm$Std).

\subsection{Performance Evaluation}
\subsubsection{Comparison with other methods}
As Table \ref{t_result_ubuntu_v1} shows, our proposed HRDE and HRDE-LTC models achieve the best performance for the Ubuntu-v1 dataset. We also find that the RDE-LTC model shows improvements from the baseline model, RDE. 

For the ubuntu-v2 dataset case, Table \ref{t_result_ubuntu_v2} reveals that the HRDE-LTC model is best for three cases (1 in 2 R@1, 1 in 10 R@2 and  1 in 10 R@5).
Comparing the same model with our implementation (RDE) and \cite{baudivs2016sentence}'s implementation (RNN), there is a large gap in the accuracy (0.610 and 0.664 of 1 in 10 R@1 for RDE and RNN, receptively). We think this is largely influenced by the data preprocessing method, because the only differences between these models is the data preprocessing, which is \cite{baudivs2016sentence}'s contribution to the research. 
We are certain that our model performs better with the exquisite datasets which adapts extensive preprocessing method, because we see improvements from the RDE model to the HRDE model and additional improvements with the LTC module in all test cases (the Ubuntu-v1/v2 and the Samsung QA).

\begin{table}[t]
\small
\centering
\begin{tabular}
{L{0.24\columnwidth}||C{0.125\columnwidth}|C{0.125\columnwidth}|C{0.125\columnwidth}|C{0.125\columnwidth}}

\hline
\multirow{2}{*}{\textbf{Model}} & \multicolumn{4}{c}{\textbf{Ubuntu-v1}} \\
& $\underset{\text{R@1}}{\text{1 in 2}}$ & $\underset{\text{R@1}}{\text{1 in 10}}$
& $\underset{\text{R@2}}{\text{1 in 10}}$ & $\underset{\text{R@5}}{\text{1 in 10}}$ \\
\hline

 \small{TF-IDF}\Tstrut~\scriptsize{[1]}
   & 0.659\Tstrut                           & 0.410\Tstrut  
   & 0.545\Tstrut   	                     & 0.708\Tstrut  \\

 \small{CNN} \scriptsize{[2]}
   & 0.848                           & 0.549
   & 0.684                           & 0.896   \\
 
 \small{LSTM} \scriptsize{[2]}
   & 0.901                           & 0.638
   & 0.784                           & 0.949   \\

 CompAgg \scriptsize{[3]}
   & 0.884                           & 0.631  
   & 0.753                           & 0.927  \\

 BiMPM \scriptsize{[4]}
   & 0.897                           & 0.665  
   & 0.786                           & 0.938  \\

\hline

 \small{RDE}\Tstrut
   & $\underset{~~\pm 0.002}{0.898}$\Tstrut & $\underset{~~\pm 0.009}{0.643}$\Tstrut
   & $\underset{~~\pm 0.007}{0.784}$\Tstrut & $\underset{~~\pm 0.002}{0.945}$\Tstrut \\
   
 \small{RDE-LTC}
   & $\underset{~~\pm 0.001}{0.903}$ & $\underset{~~\pm 0.003}{0.656}$
   & $\underset{~~\pm 0.003}{0.794}$ & $\underset{~~\pm 0.001}{0.948}$ \\
   
 \small{HRDE}
   & $\underset{~~\pm 0.001}{0.915}$ & $\underset{~~\pm 0.001}{0.681}$
   & $\underset{~~\pm 0.001}{0.820}$ & $\underset{~~\pm 0.001}{0.959}$ \\
   
 \small{HRDE-LTC}
   & $\underset{~~\pm 0.001}{\textbf{0.916}}$ & $\underset{~~\pm 0.001}{\textbf{0.684}}$ 
   & $\underset{~~\pm 0.001}{\textbf{0.822}}$ & $\underset{~~\pm 0.001}{\textbf{0.960}}$ \\
   
\hline

\hline
\end{tabular}
\caption{Model performance results for the Ubuntu-v1 dataset. Models [1-4] are from \cite{lowe2015ubuntu,kadlec2015improved,wang2016compare,wang2017bilateral}, respectively. 
}
\label{t_result_ubuntu_v1}
\end{table}
\begin{table}[t]
\small
\centering
\begin{tabular}{L{0.24\columnwidth}||C{0.125\columnwidth}|C{0.125\columnwidth}|C{0.125\columnwidth}|C{0.125\columnwidth}}
\hline
\multirow{2}{*}{\textbf{Model}} & \multicolumn{4}{c}{\textbf{Ubuntu-v2}} \\
& $\underset{\text{R@1}}{\text{1 in 2}}$ & $\underset{\text{R@1}}{\text{1 in 10}}$
& $\underset{\text{R@2}}{\text{1 in 10}}$ & $\underset{\text{R@5}}{\text{1 in 10}}$ \\
\hline

 LSTM\Tstrut~\scriptsize{[1]}
   & 0.869\Tstrut                           & 0.552\Tstrut  
   & 0.721\Tstrut                           & 0.924\Tstrut  \\

 RNN \scriptsize{[5]}
   & $\underset{~~\pm 0.002}{0.907}$ & $\underset{~~\pm 0.004}{0.664}$  
   & $\underset{~~\pm 0.004}{0.799}$ & $\underset{~~\pm 0.001}{0.951}$  \\
 
 CNN \scriptsize{[5]}
   & $\underset{~~\pm 0.003}{0.863}$ & $\underset{~~\pm 0.004}{0.587}$
   & $\underset{~~\pm 0.005}{0.721}$ & $\underset{~~\pm 0.003}{0.907}$   \\
   
 RNN-CNN \scriptsize{[5]}
   & $\underset{~~\pm 0.001}{0.911}$ & $\underset{~~\pm 0.002}{\textbf{0.672}}$
   & $\underset{~~\pm 0.002}{0.809}$ & $\underset{~~\pm 0.001}{0.956}$   \\

 $\underset{\text{(RNN-CNN)}}{\text{Attention \scriptsize{[6]}}}$
   & $\underset{~~\pm 0.002}{0.903}$ & $\underset{~~\pm 0.005}{0.653}$
   & $\underset{~~\pm 0.005}{0.788}$ & $\underset{~~\pm 0.002}{0.945}$   \\

 CompAgg \scriptsize{[3]}
   & 0.895                           & 0.641  
   & 0.776                           & 0.937  \\

 BiMPM \scriptsize{[4]}
   & 0.877                           & 0.611  
   & 0.747                           & 0.921  \\

\hline

 RDE\Tstrut 
   & $\underset{~~\pm 0.002}{0.894}$\Tstrut & $\underset{~~\pm 0.008}{0.610}$\Tstrut
   & $\underset{~~\pm 0.006}{0.776}$\Tstrut & $\underset{~~\pm 0.002}{0.947}$\Tstrut \\
   
 RDE-LTC
   & $\underset{~~\pm 0.002}{0.899}$ & $\underset{~~\pm 0.004}{0.625}$
   & $\underset{~~\pm 0.004}{0.788}$ & $\underset{~~\pm 0.001}{0.951}$ \\
   
 HRDE
   & $\underset{~~\pm 0.001}{0.914}$ & $\underset{~~\pm 0.001}{0.649}$
   & $\underset{~~\pm 0.001}{0.813}$ & $\underset{~~\pm 0.001}{0.964}$ \\
   
 HRDE-LTC
   & $\underset{~~\pm 0.002}{\textbf{0.915}}$ & $\underset{~~\pm 0.003}{0.652}$ 
   & $\underset{~~\pm 0.001}{\textbf{0.815}}$ & $\underset{~~\pm 0.001}{\textbf{0.966}}$ \\
   
\hline

\hline
\end{tabular}
\caption{Model performance results for the Ubuntu-v2 dataset. Models [1,3-6] are from \cite{lowe2015ubuntu,wang2016compare,wang2017bilateral,baudivs2016sentence,tan2015lstm}, respectively.
}
\label{t_result_ubuntu_v2}
\end{table}

In the Samsung QA case, Table \ref{t_result_samsung} indicates that the proposed RDE-LTC, HRDE, and the HRDE-LTC model show performance improvements when compared to the baseline model, TF-IDF and RDE. The average accuracy statistics are higher in the Samsung QA case when compared to the Ubuntu case. We think this is due to in the smaller vocabulary size and context variety. 
The Samsung QA dataset deals with narrower topics than in the Ubuntu dataset case. We are certain that our proposed model shows robustness in several datasets and different vocabulary  size environments.

\subsubsection{Degradation Comparison for Longer Texts}
\begin{figure*}[t]
\small
\centering
\includegraphics[width=2\columnwidth]{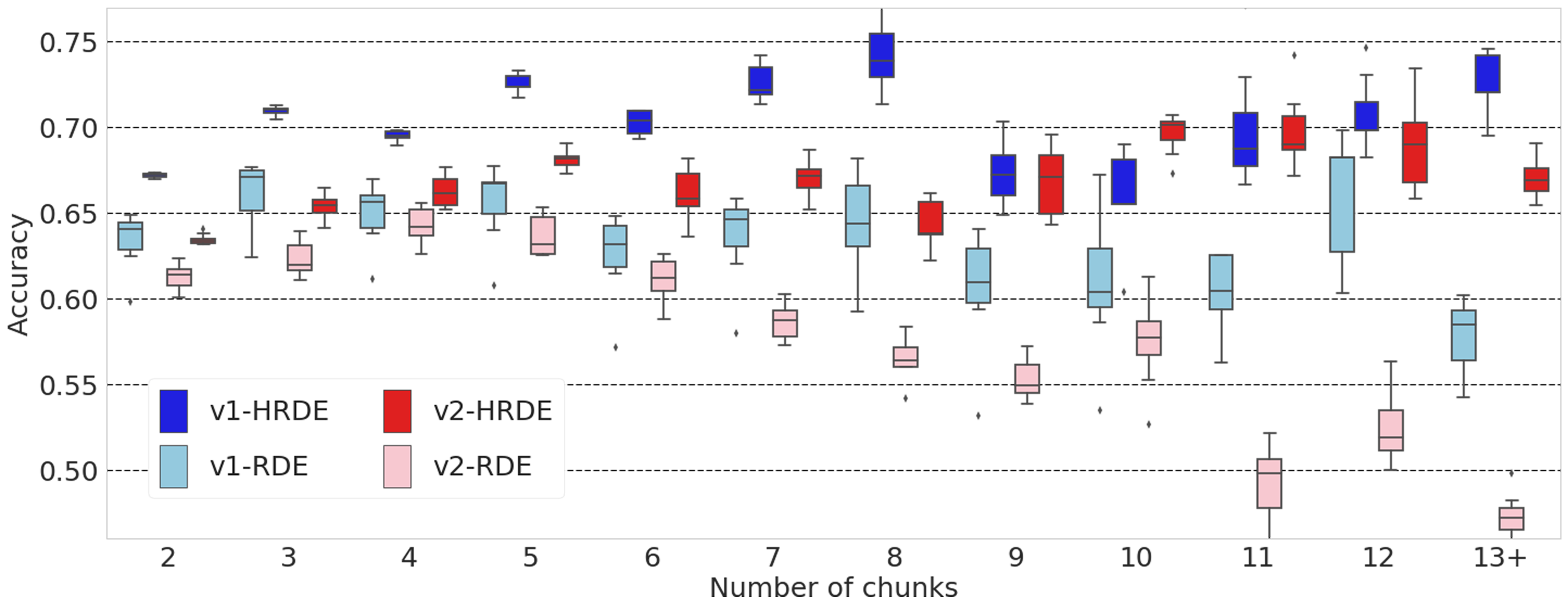}
\caption{The HRDE and RDE model performance comparisons for the number-of-chunk in the Ubuntu dataset. Each boxplot shows average accuracy with standard deviation. The HRDE models, in darker blue and red colors, show consistent performances as the number-of-chunks increased. Meanwhile, the RDE models in lighter colors show performance degradation as the number-of-chunks increased. Furthermore, 13+ indicates all data over 13-chunks.}
\label{fig_de_hrde}
\end{figure*}

To verify the HRDE model's ability compared to the baseline model RDE, we split the testset of the Ubuntu-v1/v2 datasets based on the ``number of chunks" in the \{context\}. Then, we measured the top-1 recall (same case as 1 in 10 R@1 in Table \ref{t_result_ubuntu_v1}, and \ref{t_result_ubuntu_v2}) for each group. 
Figure \ref{fig_de_hrde} demonstrates that the HRDE models, in darker blue and red colors, shows better performance than the RDE models, in lighter colors, for every ``number of chunks'' evaluations. 
In particular, the HRDE models are consistent when the ``number-of-chunks" increased, while the RDE models degrade as the ``number-of-chunks" increased.

\begin{table}[t]
\small
\centering
\begin{tabular}{L{0.21\columnwidth}||C{0.125\columnwidth}|C{0.125\columnwidth}|C{0.125\columnwidth}|C{0.125\columnwidth}}
\hline
\multirow{2}{*}{\textbf{Model}} & \multicolumn{4}{c}{\textbf{Samsung QA}} \\
& $\underset{\text{R@1}}{\text{1 in 2}}$ & $\underset{\text{R@1}}{\text{1 in 10}}$
& $\underset{\text{R@2}}{\text{1 in 10}}$ & $\underset{\text{R@5}}{\text{1 in 10}}$ \\
\hline



  TF-IDF\Tstrut~
   & 0.939\Tstrut                           & 0.834\Tstrut  
   & 0.897\Tstrut       				    & 0.953\Tstrut  \\

 RDE\Tstrut 
   & $\underset{~~\pm 0.002}{0.978}$ & $\underset{~~\pm 0.009}{0.869}$
   & $\underset{~~\pm 0.003}{0.966}$ & $\underset{~~\pm 0.001}{0.997}$ \\
   
 RDE-LTC
   & $\underset{~~\pm 0.002}{0.981}$ & $\underset{~~\pm 0.009}{0.880}$
   & $\underset{~~\pm 0.003}{0.970}$ & $\underset{~~\pm 0.001}{0.997}$ \\
   
 HRDE
   & $\underset{~~\pm 0.002}{0.981}$ & $\underset{~~\pm 0.011}{0.885}$
   & $\underset{~~\pm 0.004}{0.971}$ & $\underset{~~\pm 0.001}{0.997}$ \\
   
 HRDE-LTC
   & $\underset{~~\pm 0.002}{\textbf{0.983}}$ & $\underset{~~\pm 0.010}{\textbf{0.890}}$ 
   & $\underset{~~\pm 0.003}{\textbf{0.972}}$ & $\underset{~~\pm 0.001}{\textbf{0.998}}$ \\
   
\hline

\hline
\end{tabular}
\caption{Model performance results for the Samsung QA dataset. 
}
\label{t_result_samsung}
\end{table}
\begin{table}[t]
\small
\centering
\begin{tabular}{C{0.21\columnwidth}||C{0.18\columnwidth}|C{0.18\columnwidth}|C{0.22\columnwidth}}

\hline
\multirow{2}{*}{\textbf{\# clusters}} & \multicolumn{3}{c}{ \textbf{Accuracy (1 in 10 R@1)}} \\ 
  & Ubuntu-v1    & Ubuntu-v2    & Samsung QA    \\ \hline
1 & $\underset{~~\pm 0.009}{0.643}$ & $\underset{~~\pm 0.008}{0.610}$ & $\underset{~~\pm 0.009}{0.869}$ \\ \hline
2 & $\underset{~~\pm 0.005}{0.655}$ & $\underset{~~\pm 0.006}{0.616}$ & $\underset{~~\pm 0.011}{0.876}$ \\ \hline
3 &  $\underset{~~\pm 0.003}{\textbf{0.656}}$ &  $\underset{~~\pm 0.004}{\textbf{0.625}}$ & $\underset{~~\pm 0.010}{0.877}$ \\ \hline
4 & $\underset{~~\pm 0.005}{0.651}$ & $\underset{~~\pm 0.005}{0.622}$ & $\underset{~~\pm 0.009}{\textbf{0.880}}$ \\ \hline     

\end{tabular}
\caption{The RDE-LTC model results with different numbers of latent clusters. ``Cluster 1" is the baseline model, RDE.}
\label{t_n_topic}
\end{table}

\subsubsection{Effects of the LTC Numbers}
We analyze the RDE-LTC model for different numbers of latent clusters. 
Table \ref{t_n_topic} indicates that the model performances increase as the number of latent clusters increase (until 3 for the Ubuntu and 4 for the Samsung QA case). 
This is probably a major reason for the different number of subjects in each dataset. The Samsung QA dataset has an internal \textit{category} related to the type of consumer electronic products (6 top-level \textit{categories}; \textit{mobile}, \textit{office}, \textit{photo}, \textit{tv/video}, \textit{accessories}, and \textit{home appliance}), so that the LTC module makes clusters these \textit{categories}. The Ubuntu dataset, however, has diverse contents related to issues in using the Ubuntu system. Thus, the LTC module has fewer clusters with the sparse topic compared to the Samsung QA dataset.

\begin{figure}[t]
\small
\centering
\includegraphics[width=1\columnwidth]{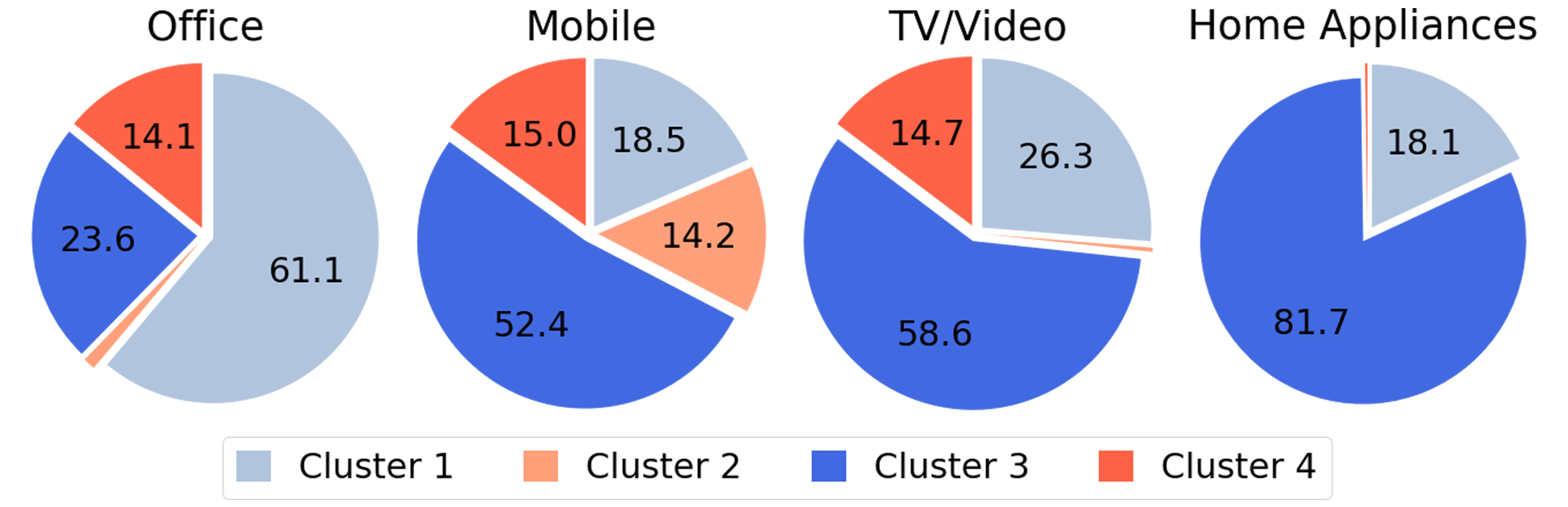}
\caption{Examples of the cluster proportions for four real \textit{categories} from $20k$ evaluated samples. Each color corresponds to each cluster.}
\label{fig_SA_topic}
\end{figure}

\subsubsection{Comprehensive Analysis of LTC}
\label{LTC_analysis}
We conduct  quantitative and qualitative analysis on the HRDE-LTC model for four latent topic clusters. 
The Samsung QA dataset has \textit{category} information; hence, latent topic clustering results can be compared with real \textit{categories}. 
We randomly choose $20k$ samples containing real \textit{category} information and evaluate each sample with the HRDE-LTC model. 
The cluster with the highest similarity among the latent topic clusters is considered a representative cluster of each sample.

\begin{table}[t]
\centering
\small

\begin{tabular}{|C{0.14\columnwidth}||L{0.72\columnwidth}|}
\hline
\textbf{Cluster} & \textbf{Example} \\ 
\hline
\hline
1 & How to adjust the brightness on the s**d300 series monitors \\
\hline
2 & How do I reject an incoming call on my Samsung Galaxy Note 3? \\
\hline
3 & How should I clean and maintain the microwave? \\
\hline
4 & How do I connnect my surround sound to this TV and what type of cables do I need \\
\hline
\end{tabular}
\caption{Example sentences for each cluster.}
\label{t_sample_topic}
\end{table}

Figure \ref{fig_SA_topic} shows proportion of four latent clusters among these samples according to real \textit{category} information. Even though the HRDE-LTC model is trained without any ground-truth \textit{category} labels, we observed that the latent cluster is formed accordingly. For instance, cluster 2 is shown mostly in ``Mobile'' \textit{category} samples while ``clusters 2 and 4'' are rarely shown in ``Home Appliance'' \textit{category} samples.

Additionally, we explore sentences with higher similarity score from the HRDE-LTC module for each four cluster. 
As can be seen in Table \ref{t_sample_topic}, ``cluster 1" contains ``screen'' related sentences (e.g., brightness, pixel, display type) while ``cluster 2" contains sentences with exclusive information related to the ``Mobile'' \textit{category} (e.g., call rejection, voice level). This qualitative analysis explains why ``cluster 2" is shown mostly in the ``Mobile'' \textit{category} in Figure \ref{fig_latent_topic}. We also discover that ``cluster 3" has the largest portion of samples. 
As ``cluster 3" contains ``security'' and ``maintenance"  related sentences (e.g., password, security, log-on, maintain), we assume that this is one of the frequently asked issues across all \textit{categories} in the Samsung QA dataset.  Table \ref{t_sample_topic} shows example sentences with high scores from each cluster.

\section{Conclusion}
\label{coclusion}
In this paper, we proposed the HRDE model and LTC module. HRDE showed higher performances in ranking answer candidates and less performance degradations when dealing with longer texts compared to conventional models. 
The LTC module provided additional performance improvements when combined with both RDE and HRDE models, as it added latent topic cluster information according to dataset properties.
With this proposed model, we achieved state-of-the-art performances in Ubuntu datasets.
We also evaluated our model in real world question answering dataset, Samsung QA. This demonstrated the robustness of the proposed model with the best results.

\section*{Acknowledgments}
K. Jung is with the Department of Electrical and Computer Engineering, ASRI, Seoul National University, Seoul, Korea. This work was supported by Basic Science Research Program through the National Research Foundation of Korea (NRF) funded by the Ministry of Education (NRF-2016M3C4A7952587), the Ministry of Trade, Industry \& Energy (MOTIE, Korea) under Industrial Technology Innovation Program (No.10073144).

\bibliography{naaclhlt2018}
\bibliographystyle{acl_natbib}

\appendix
\section{More examples of the dataset}
\label{supplemental}

\subsection{Ubuntu dataset}
\begin{table}[hb]
\centering

\begin{tabular}{|L{0.965\columnwidth}|}

\hline
\textbf{Question} \Tstrut \\ 
\hline
``what will happend if i unmounted the ubuntu partition", ``it will unmount , unless it is in use", ``srr i did n't got it" 
\Tstrut\\
\hline
\end{tabular}

\vskip 2mm

\begin{tabular}{|L{0.82\columnwidth}|C{0.09\columnwidth}|}
\hline
\textbf{Answer} & \textit{flag} \Tstrut \\ 
\hline
``you cannot unmount a partition if it is currently in use" & 1
\Tstrut \\  \hline

``why do you not have a backup if the data is important ?" & 0
\Tstrut \\ \hline

\end{tabular}

\caption{Example of the Ubuntu-v2 dataset.}
\label{appendix_ubuntu}
\end{table}


\subsection{Samsung QA dataset}
\begin{table}[hb]
\centering

\begin{tabular}{|L{0.965\columnwidth}|}

\hline
\textbf{Question} \Tstrut \\ 
\hline
how can i place the current call on hold at any point during a conversation ?
\Tstrut\\
\hline
\end{tabular}

\vskip 2mm

\begin{tabular}{|L{0.82\columnwidth}|C{0.09\columnwidth}|}
\hline
\textbf{Answer} & \textit{flag} \Tstrut \\ 
\hline
you can place the current call on hold at any point during a conversation . you can also make another call while you have a call in progress if your network supports this service . 1 while on a call , tap hold . this action places the current caller on hold . 2 you can later reactivate this call by tapping unhold . & 1
\Tstrut \\  \hline

please try to do a soft reset . turn of the phone , remove and put the battery back after 1-2 minutes . we also recommend you to clear the data of the samsung keyboard . 1 from the home screen , touch application 2 select settings 3 select application manager 4 touch the all tab 5 select samsung keyboard 6 tap on clear data . & 0
\Tstrut \\ \hline

\end{tabular}

\caption{Example of the Samsung QA dataset.}
\label{appendix_samsungQA}
\end{table}


\end{document}